%% file: emnlp2022.tex
\pdfoutput=1

\documentclass[11pt]{article}

\usepackage{EMNLP2022}

\usepackage{times}
\usepackage{latexsym}

\usepackage[T1]{fontenc}

\usepackage[utf8]{inputenc}

\usepackage{microtype}

\usepackage{inconsolata}

\usepackage{amsmath}
\usepackage{amssymb}
\usepackage{booktabs}
\usepackage{multirow}
\usepackage{xcolor}
\usepackage{graphicx}

\title{UniRel: Unified Representation and Interaction for Joint Relational Triple Extraction}

\author {
    Wei Tang \textsuperscript{\rm 1,2},
    Benfeng Xu \textsuperscript{\rm 1,2},
    Yuyue Zhao \textsuperscript{\rm 1,2},
    Zhendong Mao \textsuperscript{\rm 1,2},
    Yifeng Liu \textsuperscript{\rm 3} \\
    \textbf{Yong Liao} \textsuperscript{\rm 1,2} \thanks{\quad Corresponding author.} , 
    \textbf{Haiyong Xie} \textsuperscript{\rm 1,2}  \\
    \textsuperscript{\rm 1} University of Science and Technology of China, Anhui, China\\
    \textsuperscript{\rm 2} CCCD Key Lab of Ministry of Culture and Tourism, Anhui, China \\
    \textsuperscript{\rm 3} National Engineering Laboratory for Risk Perception and Prevention (RPP), Beijing, China\\
    \{weitang, benfeng, yyzha0\}@mail.ustc.edu.cn, zdmao@ustc.edu.cn, liuyifeng3@cetc.com.cn \\
    \{yliao, hxie\}@ustc.edu.cn
}

\begin{document}
\maketitle
\begin{abstract}
Relational triple extraction is challenging for its difficulty in capturing rich correlations between entities and relations.
Existing works suffer from 1) heterogeneous representations of entities and relations, and 2) heterogeneous modeling of entity-entity interactions and entity-relation interactions. Therefore, the rich correlations are not fully exploited by existing works. 
In this paper, we propose UniRel to address these challenges.
Specifically, we unify the representations of entities and relations by jointly encoding them within a concatenated natural language sequence, and unify the modeling of interactions with a proposed Interaction Map, which is built upon the off-the-shelf self-attention mechanism within any Transformer block.
With comprehensive experiments on two popular relational triple extraction datasets, we demonstrate that UniRel is more effective and computationally efficient.
The source code is available at https://github.com/wtangdev/UniRel.
\end{abstract}

\input{introduction.tex}

\input{related_work.tex}

\input{methodology.tex}

\input{expeiments.tex}

\input{analysis.tex}

\section*{Limitations}
There are two limitations we want to discuss in this section:
\begin{itemize}
    \item First, for clarity, we select the corresponding words for each relation in a manual way, which would be sophisticated for schema with much relations. We will next to try to design an auto-verbalizer for relations.
    \item Second, the evaluation benchmarks (NYT and WebNLG) for joint relational triple extraction are produced with much annotated training data, which is expensive for real application. The performance of our model in low-resource scenario needs to be validated. However, existing benchmarks for low-resource \cite{han-etal-2018-fewrel, gao-etal-2019-fewrel} are limited to the simple scenario of sentence-level relation classification. We would like to explore the the idea of unified representation and unified interaction towards joint RTE in low resource scenario in the future. 
\end{itemize}

\section*{Acknowledgements}

This work is supported by the National Key Research and Development Program of China (2021YFC3300500).

\bibliography{anthology,custom, joint-rte}
\bibliographystyle{acl_natbib}

\appendix

\input{appendix.tex}

\end{document}

%% file: introduction.tex
\section{Introduction}

Relational Triple Extraction (RTE) aims to identify entities and their semantic relations jointly. It extracts structured triples in the form of\textit{ <subject-relation-object>} from raw texts in an end-to-end manner, and is a crucial task towards automatically constructing large-scale knowledge bases  \cite{DBLP:journals/corr/abs-2103-16929}.

Early works try to solve RTE tasks in a pipelined fashion involving two sub-tasks \cite{zelenko_kernel_nodate, chan-roth-2011-exploiting}, where entities are firstly recognized, and relations are then assigned for each extracted entity pair. Therefore, such methods fail to capture the implicit correlation between these two isolated sub-tasks and are thus prone to propagated errors \cite{li-ji-2014-incremental}. 

Instead, many researchers \cite{miwa-bansal-2016-end, zheng-etal-2017-joint, zeng-etal-2018-extracting} seek to jointly extract <s-r-o> triples in an end-to-end manner. 
For example, \citet{wei-etal-2020-novel} propose a cascaded network that identifies subjects first and then recognizes corresponding objects for relations. \citet{DBLP:conf/acl/ZhengWCYZZZQMZ20} decompose RTE into three sub-tasks. \citet{wang_tplinker_2020} and \citet{Shang2022OneRelJE} extract relational triples in one stage to eliminate exposure bias.
In general, these works accomplish end-to-end extraction by various ways to factorize and re-assemble the label space of relational triples.
However, very often, there exist rich informative correlations between entities and relations, and these correlations can hardly be captured by the superficial constraints applied in label space only.
Specifically, existing works fall short of two aspects: 1) heterogeneous representations between entities and relations; 2) ignorance of interactive dependencies between entity-level interactions and relation-level interactions.

\begin{figure}[t]
  \centering
  \includegraphics[width=0.95\columnwidth]{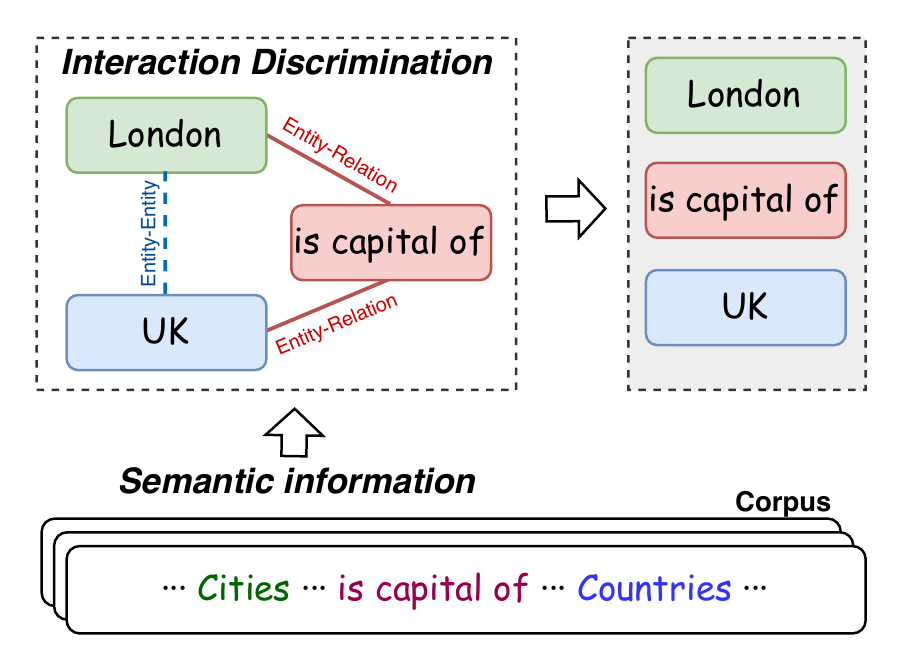} 
  \caption{We leverage semantic information to unify the representation of entities and relations. Relational triples are extracted by modeling the entity-entity interaction (blue dashed line) and entity-relation interaction (red solid line) in a unified way.}
  \label{fig:model_intro}
\end{figure}

For representations, existing works mainly focus on how to better capture the contextual information of entities, while ignoring the equally important semantic meaning of relations.
Generally, relations are simply represented as atomic label ids indicating a specific dimension in the newly initialized classifier (\citealp{wei-etal-2020-novel, wang_tplinker_2020, DBLP:conf/acl/ZhengWCYZZZQMZ20}), resulting in its heterogeneity with language model augmented entity representations.
Such heterogeneity prevents models from capturing the intrinsic correlations between entities and relations in the semantic space.
For instance, from the semantic meaning of the relation \texttt{is\_capital\_of}, we can infer that triples involving are related to locations: the subject is supposed to be a city, while the object should be a country.
We argue that it is important to build unified representations for both entities and relations.

For interactions, we exemplify the interdependence between the entity-entity interactions and entity-relation interactions in Figure \ref{fig:model_intro}.
We can easily determine that \textit{London} and \textit{UK} are correlated in the prerequisite of given the interactions of (\textit{London}-\texttt{is\_capital\_of}) and (\textit{UK}-\texttt{is\_capital\_of}).
However, existing works either enumerate all entity-relation-entity triples (\citealp{wang_tplinker_2020, Shang2022OneRelJE}), which suffer from huge prediction space, or model them with separate modules (\citealp{wei-etal-2020-novel, DBLP:conf/acl/ZhengWCYZZZQMZ20, li-etal-2021-tdeer}) respectively for named entity recognition (NER) and relation extraction (RE), resulting in neglect of interdependencies. 
In general, the absence of unified modeling of interactions limits these methods from fully utilizing their interdependencies for better extraction.

In this paper, we propose \textbf{UniRel} with \textbf{\textit{Unified Representation}} and \textbf{\textit{Interaction}} to resolve both the heterogeneity of representations and the absence of interaction dependencies.
We first encode both relation and entity into meaningful sequence embeddings to construct unified representations.
Based on the semantic definition, candidate relations are first converted to natural language texts, and form a consecutive sequence together with the input sentence.
We then apply a Transformer-based Pre-trained Language Model (PLM) to encode the sequence while intrinsically capturing their informative correlations.
We then propose a novel solution for \textbf{\textit{Unified Interaction}}, where we simultaneously model the entity-entity interactions and entity-relation interactions in one single \texttt{Interaction Map} by leveraging the off-the-shelf self-attention mechanism inside Transformer.
Besides, benefiting from the design of Interaction Map, UniRel preserves the advantages for end-to-end extraction and is superior in computational efficiency.

We conduct comprehensive experiments on two popular datasets: NYT \cite{DBLP:conf/pkdd/RiedelYM10} and WebNLG \cite{DBLP:conf/acl/GardentSNP17}, and achieve a new state-of-the-art. We summarize our contributions as follows:
\begin{itemize}
  \item We propose unified representations of entities and relations by jointly encoding them within a concatenated natural language sequence, which fully exploits their contextualized correlations and leverages the semantic knowledge learned from LM pre-training.
  \item We propose unified interactions to capture the interdependencies between entity-entity interactions and entity-relation interactions. This is innovatively achieved by the proposed Interaction Map built upon the off-the-shelf self-attention mechanism within any Transformer block.
   \item We show that UniRel achieves the new state-of-the-art for RTE tasks, while preserving the superiority of being end-to-end and computationally efficient.
\end{itemize}

%% file: related_work.tex
\section{Related Work}

Early works \cite{zelenko_kernel_nodate, chan-roth-2011-exploiting} apply pipeline approaches that divide relational triple extraction into two isolated sub-tasks: first do named entity recognition to
extract all entities, and then apply relation extraction to identify relations for each entity pair. 
However, these methods always suffer from error propagation for failing to
capture the implicit correlation between these two isolated sub-tasks. 

To tackle these issues, recent researches focus on jointly extracting entities and relations. Previous feature-based joint models \cite{DBLP:conf/coling/YuL10,DBLP:conf/emnlp/MiwaS14,li-ji-2014-incremental,DBLP:conf/www/RenWHQVJAH17} require complex feature engineering and heavily depend on NLP tools.
Researchers propose neural network-based joint models to eliminate hand-craft features. \citeauthor{miwa-bansal-2016-end} \shortcite{miwa-bansal-2016-end} propose a model to jointly learn entities and relations through parameter sharing.
\citeauthor{zheng-etal-2017-joint} \shortcite{zheng-etal-2017-joint} transform RTE into a sequence tagging problem, which unifies the annotation role of entities and relations.

Despite their success, most models cannot deal with complex scenarios where one sentence consists of multiple overlapping relational triples sharing a single entity (SingleEntityOverlap, SEO) or an entity pair (EntityPairOverlap, EPO).  
To handle the problem, researchers (\citealp{zeng-etal-2018-extracting, zeng-etal-2019-learning, DBLP:conf/aaai/NayakN20, DBLP:conf/aaai/YeZDCTHC21}) propose generative models that view triple as a token sequence.
Some works (\citealp{wang_tplinker_2020, ren-etal-2021-novel, Shang2022OneRelJE}) introduce methods to extract triples in one-stage but suffer from huge prediction space.
Other researchers (\citealp{wei-etal-2020-novel, yuan_relation-specific_2020, zheng-etal-2021-prgc, li-etal-2021-tdeer, wu-shi-2021-synchronous}) decompose RTE into different sub-tasks, but learn the interaction between sub-tasks only by input sharing, or falling into the cascade error. PFN \cite{yan-etal-2021-partition} proposes a partition filter network to fuse the task representation of NER and RE, but still models entity-entity interactions and entity-relation interactions with separate modules. In this work, UniRel unifies the modeling of the two kinds of interactions in one single Interaction Map to fully capture their interdependencies and is superior in computational efficiency.

More recently, \citet{xu-etal-2022-emrel} propose EmRel that explicitly introduce relation representation to leverage the rich interactions across relations, entities, and context. However, it still suffers from heterogeneity between entities and the newly initialized embeddings of relations.
Some approaches \cite{han2021ptr, chen-li-2021-zs, 10.1145/3485447.3511998} introduce prompt-tuning to extract relation with semantic information. They transform the relation extraction task into a masked language modeling problem. However, such methods focus on the simple scenario of sentence-level relation classification without capturing the correlations between entities and relations. In terms of technical designs, SSAN \cite{Xu_Wang_Lyu_Zhu_Mao_2021} also delves into the self-attention layer within the Transformer to model structural interaction, but it makes extra adaptions for standard self-attention mechanism and focuses on document-level RE tasks. Compared to these approaches, our work aims to extract relational triples in complex scenarios where rich intrinsic correlations exist between entities and relations. In this paper, we unify the representations and interactions to fully exploit the correlations to extract entities and relations jointly.

%% file: methodology.tex
\section{Methodology}

\begin{figure*}[t]
    \centering
    \includegraphics[width=2.0\columnwidth]{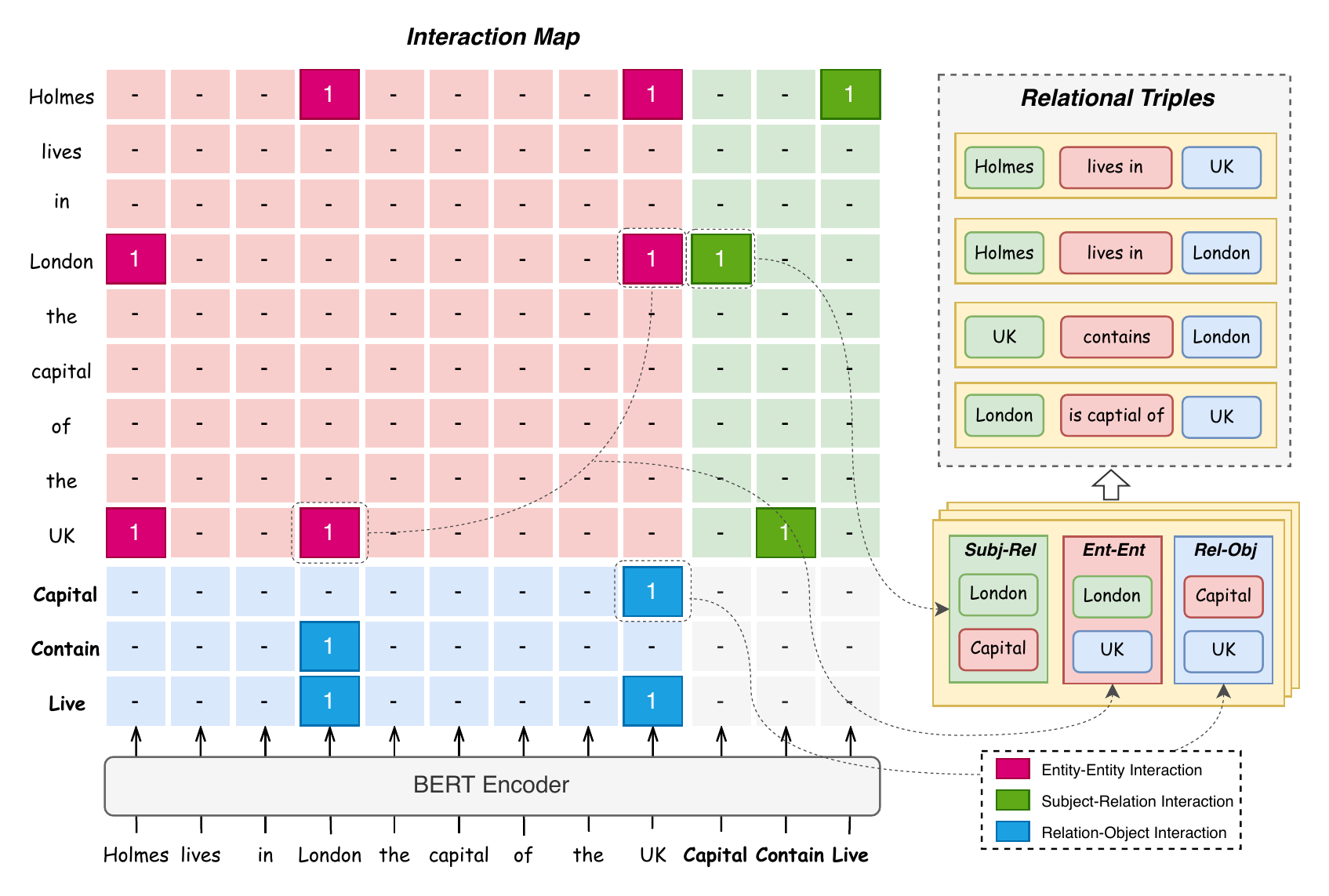} 
    \caption{We input the concatenation of the input sentence and the natural language texts of relations (in bold). The Interaction Map is learned from the attention map inside the 12th layer of BERT Encoder, which consists of Entity-Entity Interaction (red rectangle) and Entity-Relation Interaction (green rectangle for subject and blue rectangle for object). Relational triples are extracted intuitively from the map.}
    \label{fig:model_structure}
  \end{figure*}

In this section, we present our model in detail. We first introduce the problem formulation in Section \ref{sec:pf}. Then, we introduce the Unified Representation and the Unified Interaction in Section \ref{sec:up} and Section \ref{sec:ui}, respectively. Finally, we present the details of training and decoding in Section \ref{sec:td}.

\subsection{Problem Formulation} \label{sec:pf}

Given a sentence $X = \{x_1, x_2, \cdots, x_N\}$ with $N$ tokens, the goal of joint relational triple extraction is to identify all possible triples $T = [(s_l, r_l, o_l)]_{l=1}^L$ from $X$, where $s_l$, $r_l$, $p_l$ represent the subject, the object, and their relation, respectively, and $L$ is the number of triples.
The subject and object are entity mentions $E = \{e_1, e_2, \cdots, e_k\}$ from sentence $X$,
k is the number of entities.
The relation is from pre-defined relation set $R = \{R_1, R_2, \cdots, R_M\}$ with $M$ types.
Note that entities and relations might be shared among triples.

\subsection{Unified Representation} \label{sec:up}

We first convert relations in the schema to natural language texts to represent them in the same form as the input sentence.
For clarity, the conversion is performed through a verbalizer with human-picked words. The relation word is basically the most informative word within the label name that preserves its semantics, for example, ``founders'' for relation ``\texttt{/business/company/founders}''. 
We then input the concatenation of the input sentence and the natural language texts of relations to a Transformer-based PLM encoder.
We use BERT \cite{DBLP:conf/naacl/DevlinCLT19} as the PLM in this work. 
The inputs are then transferred to a sequence of input embeddings by searching the corresponding ids from the embedding table. 
\begin{equation}
    T = \text{Concat}(T_s, T_p)
\end{equation}
\begin{equation}
    H = E[T]
\end{equation}
where $H \in \mathbb{R}^{(N+M)\times d_h}$ is the input embedding vector. $d_h$ is the embedding size. $T_s$ and $T_p$ are the input ids of the input sentence and the relations, respectively. $E$ is the embedding table in BERT.

After obtaining the input embeddings, the encoder captures the correlations between each input word with the self-attention mechanism.
To be specific, Transformer-based PLMs comprise stacked Transformer \cite{DBLP:conf/nips/VaswaniSPUJGKP17} layers consisting of multiple attention heads. Each head applies three separate linear transformations to transform the input embeddings $H$ to query, key, and value vectors $Q$, $K$, $V$, and then computes the attention weights between all pairs of words by Softmax-normalized dot production of $Q$ and $K$, which then is fused with $V$ as follows:
\begin{equation}
    \text{Attention}(Q, K, V) = \text{softmax}(\frac{QK^T}{\sqrt{d_h}})V.
\end{equation}

Each Transformer layer generates token embeddings from the previous layer's output with the self-attention mechanism. We denote the $H_i$ as the output of the $i$-the Transformer layer.
As we take both entities and relations into the input embedding $H$, the $H_i$, encoded by such a deep Transformer network, fully captures the rich intrinsic correlations between entities and relations. 
The above steps make the representations of entities and relations unified into one semantic embedding vector $H_i$ with rich correlational information.

\subsection{Unified Interaction} \label{sec:ui}

The interactions between the triple elements, \textit{Entity-Entity Interaction} and \textit{Entity-Relation Interaction}, can be directly used to extract relational triples. As shown in Figure \ref{fig:model_structure}, Triple (\textit{London}, \texttt{is\_capital\_of}, \textit{UK}) can be determined if knowing the interactions between (\textit{London}-\textit{UK}), (\textit{London}-\texttt{is\_capital\_of}), and (\textit{UK}-\texttt{is\_capital\_of}). Motivated by this, we design an Interaction Map to model the two kinds of interactions simultaneously.

\subsubsection{Entity-Entity Interaction}

Entity-entity interaction is defined for identifying entity pairs that can be used to form valid relational triples.
Given two entities $e_a$ and $e_b$ from sentence $X$, we regard entity pair $(e_a, e_b)$ are interacted only when there exists relation $r$ that can be formed as valid triples together with them, and both $(e_a, r, e_b)$ and $(e_b, r, e_a)$ are allowed.
For example, in Figure \ref{fig:model_structure}, (\textit{Holmes} - \textit{London}), as well as (\textit{London} - \textit{Holmes}), are supposed to be interacted for the existing triple (\textit{Holmes}, \texttt{lives in}, \textit{London}), while unrelated for (\textit{Holmes} - \textit{capital}) since no valid triple consists of them.
Formally, we define the entity-entity interaction indicator function $I_e(\cdot)$ as follows:
\begin{equation}
    I_e(e_a, e_b) =
    \begin{cases}
        \textit{True} &(e_a, r, e_b) \in T~ \text{or} \\&(e_b, r, e_a) \in T, \exists r \in R\\
        \textit{False} &\text{otherwise} \\
    \end{cases},
\end{equation}
$I_e(e_b, e_a) = I_e(e_a, e_b)$, as entity-entity interaction is symmetrical.

\subsubsection{Entity-Relation Interaction}

Entity-relation interaction recognizes correlated entities for each relation.
Given a relation $r$, we regard entity $e$ as interacting with $r$ when existing triples consisting of $e$ as either subject or object and $r$ as the relation.
As the relation is directional, we define entity-relation interaction asymmetrically to distinguish subject entities and object entities, as shown in the upper right part (Subject-Relation) and lower left part (Relation-Object) of the map in Figure \ref{fig:model_structure}, respectively.
For instance, the interaction value of (\textit{London} - \texttt{is\_capital\_of}) is supposed to be \textit{True} because of the valid triple (\textit{London}, \texttt{is\_capital\_of}, \textit{UK}), while \textit{False} for (\textit{UK} - \texttt{is\_capital\_of}) since it is impossible for \textit{UK} to be the subject of the relation \texttt{is\_capital\_of}.
We formally define the indicator function $I_r(\cdot)$ of entity-relation interaction as follows:
\begin{equation}
    I_r(e, r) =
    \begin{cases}
        \textit{True} &(e, r, o) \in T, \exists o \in E\\
        \textit{False} &\text{otherwise} \\
    \end{cases},
\end{equation}
\begin{equation}
    I_r(r, e) =
    \begin{cases}
        \textit{True} &(s, r, e) \in T, \exists s \in E\\
        \textit{False} &\text{otherwise} \\
    \end{cases},
\end{equation}
where $I_r(e, r)$ and $I_r(r, e)$ are defined for identifying entity $e$ as subject and object for relation $r$, respectively.

\subsubsection{Interaction Discrimination}

Transformer layers bring powerful deep correlation capturing ability to BERT. Therefore, As shown in Figure \ref{fig:model_structure}, we comprise the two kinds of interactions into one single Interaction Map, which is in the same form as the attention map computed by the Transformer layer. Then we directly take the last Transformer layer of BERT for Interaction Discrimination. As the Interaction Map is not restricted to a normalized matrix, after obtaining $Q$, $K$ from $H_{11}$, the embeddings generated by the last layer of BERT, we average the dot production of $Q$ and $K$ of all heads and directly apply the sigmoid function to obtain the results. The detailed operations are as follows:
\begin{equation}
    \mathbf{I} = \text{sigmoid}(\frac{1}{T}\sum_t^T{\frac{Q_tK^T_t}{\sqrt{d_h}}}),
\end{equation}
where $\mathbf{I} \in \mathbb{R}^{(N+M) (N+M)}$ is the interaction matrix corresponding to the Interaction Map. $T$ is the number of heads. $W_t^Q$ and $W_t^K$ are trainable weights.  We consider $\mathbf{I}(\cdot)$ valid when the value of $\mathbf{I}(\cdot)$ exceeds threshold $\sigma$.

UniRel captures the interactive dependencies as the entity-entity interactions and entity-relation interactions are simultaneously modeled in one single Interaction Map. Besides, with Unified Interaction, the prediction space is narrowed down to $O((N+M)^2)$, which is much smaller than the most recent work, OneRel, which predicts triples in the complexity of $O(N \times M \times N)$.

\subsection{Training and Decoding} \label{sec:td}
The binary cross entropy loss is used for training:
\begin{equation}
    \begin{aligned}
        \mathcal{L} = -\frac{1}{(N+M)^2} \sum_{i}^{N+M}\sum_{j}^{N+M} (\mathbf{I^*}_{i,j}\log\mathbf{I}_{i,j}\\
        + (1-\mathbf{I^*}_{i,j})\log(1-\mathbf{I}_{i,j})),
    \end{aligned}
\end{equation}
where $\mathbf{I}^*$ is the ground truth matrix of the Interaction Map.

For decoding, we first recognize all valid subject entities and object entities for each relation from entity-relation interactions $I_r$ (the green box and blue box in the lower right part of Figure \ref{fig:model_structure}).
Then we enumerate all candidate entity pairs for each relation pruned by entity-entity interactions $I_e$ (the red box in the lower right part of Figure \ref{fig:model_structure}).

Such a decoding method can address the complex scenarios with overlapping patterns containing \textit{EntityPairOverlap} (EPO) and \textit{SingleEntityOverlap} (SEO) for all arrangements of the entities and relations are taken into account.
As shown in Figure \ref{fig:model_structure}, the extracted triples contain SEO triples: (\textit{Holmes}, \texttt{lives\_in}, \textit{UK}) and (\textit{Holmes}, \texttt{lives\_in}, \textit{London}), and the EPO triples: (\textit{UK}, \texttt{contains}, \textit{London}) and (\textit{London}, \texttt{is\_capital\_of}, \textit{UK}).

%% file: expeiments.tex
\section{Experiments}

\begin{table*}[h]
    \centering
    \begin{tabular}{lccccccccc}
        \toprule
        \multirow{2}*{Model}  & \multicolumn{3}{c}{NYT} & \multicolumn{1}{c}{~} & \multicolumn{3}{c}{WebNLG}   \\
        \cmidrule(lr){2-4} \cmidrule(lr){6-8}
                                          & Prec.  & Rec.   & F1 &    & Prec.  & Rec.   & F1  \\
        \midrule 
        NovelTagging \cite{zheng-etal-2017-joint}  & 62.4  & 31.7 & 42.0 &  & 52.5 & 19.3 & 28.3      \\
        CopyRE \cite{zeng-etal-2018-extracting}    & 61.0  & 56.6 & 58.7 &  & 37.7   & 36.4   & 37.1     \\
        GraphRel \cite{fu-etal-2019-graphrel}      & 63.9  & 60.0 & 61.9 &  & 44.7   & 41.1   & 42.9            \\
        OrderCopyRE \cite{zeng-etal-2019-learning} & 77.9  & 67.2 & 72.1 &  & 63.3   & 59.9   & 61.6        \\
        $\text{CasRel}_\textit{BERT}$ \cite{wei-etal-2020-novel}    & 89.7   & 89.5   & 89.6 &  & 93.4   & 90.1   & 91.8     \\
        $\text{TPlinker}_\textit{BERT}$ \cite{wang_tplinker_2020}   & 91.3   & 92.5   & 91.9 &  & 91.7   & 92.0   & 91.9    \\        
        $\text{PRGC}_\textit{BERT}$ \cite{DBLP:conf/acl/ZhengWCYZZZQMZ20}      & 93.3   & 91.9   & 92.6 &   & 94.0   & 92.1   & 93.0    \\
        $\text{R-BPtrNet}_\textit{BERT}$ \cite{chen_jointly_2021} & 92.7 & 92.5 & 92.6 & & 93.7 & 92.8 & 93.3 \\
        PFN \cite{yan-etal-2021-partition} & - & - & 92.4 & & - & - & 93.6  \\ 
        $\text{TDEER}_{\textit{BERT}}$ \cite{li-etal-2021-tdeer} & 93.0 & 92.1 & 92.5 &  & 93.8 & 92.4 & 93.1  \\
        $\text{GRTE}_{\textit{BERT}}$ \cite{ren-etal-2021-novel} & 92.9 & 93.1 & 93.0 &  & 93.7 & 94.2 & 93.9 \\
        $\text{EmRel}$ \cite{xu-etal-2022-emrel} & 91.7 & 92.5 & 92.1 & & 92.7 & 93.0 & 92.9 \\
        $\text{OneRel}_\textit{BERT}$ \cite{Shang2022OneRelJE} & 92.8 & 92.9 & 92.8 &  & 94.1 & 94.4 & 94.3  \\
        \midrule                          
        $\text{UniRel}$ & \textbf{93.5} & \textbf{94.0} & \textbf{93.7} & &  \textbf{94.8} & \textbf{94.6} & \textbf{94.7} \\
        $\text{UniRel}_{\textit{unused}}$ & 93.1 & 93.2 & 93.1 & & 75.1 & 68.6 & 71.7 \\
        $\text{UniRel}_{\textit{separate}}$ & 92.6 & 93.7 & 93.1 & & 93.5 & 94.4 & 93.9 \\
        \bottomrule
    \end{tabular}
    \caption{\label{tab:main_results}Main results. The highest scores are in bold.}
\end{table*}

\subsection{Datasets and Evaluation} \label{sec:de}

We evaluate the proposed method on two widely used benchmark datasets NYT \cite{DBLP:conf/pkdd/RiedelYM10} and WebNLG \cite{DBLP:conf/acl/GardentSNP17}.
NYT dataset is produced by distant supervision from New York Times articles which has 24 predefined relations.
WebNLG was first created for natural language generation task and is adapted to relational triple extraction by \citet{zeng-etal-2018-extracting}, which has 171 predefined relations.
The statistics of the datasets are shown in Table \ref{tab:datasets}. 
We evaluate our method on standard data splitting and report the standard micro Precision (Prec.), Recall (Rec.), and F1-score on test set following the same setting as \citet{zeng-etal-2019-learning}. 

Our model is implemented based on Pytorch. We optimize the parameters by Adam \cite{DBLP:journals/corr/KingmaB14} using learning rates 3e-5 and 5e-5 for NYT and WebNLG, respectively. The learning rates are searched in \{3e-5, 5e-5, 7e-5\}.
We also conduct weight decay \cite{DBLP:conf/iclr/LoshchilovH19} with a rate of 0.01.
The batch size is 24/6 for NYT/WebNLG and trained for 100 epochs. We use cased BERT-base \footnote{https://huggingface.co/bert-base-cased} with 108M parameters as the PLM and set the max length of the input sentence to 100 to keep in line with previous work.
The size of attention head $d_h$ is 64. The threshold $\sigma$ is set as 0.5. We tune the parameters on the development set. Our experiments are conducted on one NVIDIA V100 32GB GPU.

\begin{table}[h]
    \setlength\tabcolsep{2.3pt}
    \begin{tabular}{lccccccc}
        \toprule
        \multirow{2}*{Dataset} & \multirow{2}*{Train} & \multirow{2}*{Test} & \multicolumn{4}{c}{Overlapping Pattern}\\
        \cmidrule(lr){4-7}
         &  &  & Normal & SEO & EPO & SOO \\
            
        \midrule
        NYT    & 56195   & 5000  & 3266 & 1297 & 978 & 45 \\
        WebNLG & 5019    & 703   & 245  & 457  & 26  & 84\\
        \bottomrule
    \end{tabular}
    \caption{Statistics of evaluation datasets. Overlapping patterns are counted on test set.}
    \label{tab:datasets}
\end{table}

\subsection{Results} \label{sec:res}

For comparison, We employed twelve strong models as baselines consisting of the SOTA models PRGC \cite{DBLP:conf/acl/ZhengWCYZZZQMZ20}, PFN \cite{yan-etal-2021-partition}, TDEER \cite{li-etal-2021-tdeer}, GRTE \cite{ren-etal-2021-novel} and OneRel \cite{Shang2022OneRelJE}. We take the experiment results from the original papers of these baselines directly.

Table \ref{tab:main_results} shows the results of our model against other baseline methods on all datasets.
Many previous baselines achieve F1-score of over 90\% on both datasets, especially on WebNLG, which already exceed human-level performance. UniRel achieves +0.7\% and +0.4\% improvements over F1-scores on NYT and WebNLG and outperforms all the baselines in terms of all the evaluation metrics, which shows the superiority of our model.

\begin{table*}[t]
    \renewcommand{\arraystretch}{1}
    \setlength\tabcolsep{2pt}
    \centering
    \begin{tabular}[]{lccccccccc}
        \toprule
        Model        & Normal & SEO  & EPO   & SOO   & L$ = 1$   & L$ = 2$   & L$ = 3$   & L$ = 4$   & L$ \geq 5$  \\
        \midrule        
        CasRel                              & 87.3   & 91.4 & 92.0  & 77.0  & 88.2      & 90.3      & 91.9      & 94.2       & 83.7       \\
        TPlinker                           & 90.1   & 93.4 & 94.0  & 90.1  & 90.0      & 92.8      & 93.1      & 96.1       & 90.0       \\
        PRGC    & 91.0   & 94.0 & 94.5  & 81.8  & 91.1   & 93.0    & 93.5   & 95.5    & 93.0       \\
        R-BPtrNet & 90.4 & 94.4 & \textbf{95.2} & - & 89.5 & 93.1 & 93.5 & \textbf{96.7} & 91.3 \\
        PFN & 90.2 & \textbf{95.3} & 94.1 & - & 90.5 & 92.9 & 93.7 & 96.3 & 92.6 \\
        TDEER & 90.8 & 94.1 & 94.5 & - & 90.8 & 92.8 & 94.1 & 95.9 & 92.8 \\
        GRTE & 91.1 & 94.4 & 95.0 & - & 90.8 & 93.7 & 94.4 & 96.2 & 93.4 \\
        OneRel & 90.6 & 95.1 & 94.8 & \textbf{90.8} & 90.5 & 93.4 & 93.9 & 96.5 & \textbf{94.2} \\
        \midrule        
        UniRel   & $\textbf{91.6}_{\pm 0.3}$  & $\textbf{95.3}_{\pm 0.2}$ & $\textbf{95.2}_{\pm 0.1}$  & $89.8_{\pm 3.6}$  & $\textbf{91.5}_{\pm 0.3}$  & $\textbf{94.3}_{\pm 0.2}$   & $\textbf{94.5}_{\pm 0.3}$  & $96.6_{\pm 0.2}$  & $\textbf{94.2}_{\pm 0.8}$      \\
        
        \bottomrule
    \end{tabular}
    \caption{F1-score on sentences with different overlapping patterns and different triple numbers. L is the number of triples in one sentence. All the compared models are implemented with BERT. We report the average results of UniRel  of the five runs with different random seeds. The highest scores are in bold.}
    \label{tab:complex_results}
    
\end{table*}

To further study the ability to handle the overlapping problem and extracting multiple triples, following previous works \cite{wei-etal-2020-novel,wang_tplinker_2020,DBLP:conf/acl/ZhengWCYZZZQMZ20, Shang2022OneRelJE}, we conduct further experiments on different types subsets of NYT.

As shown in Table \ref{tab:complex_results}, the results indicate the effectiveness of our model in complex scenarios. Our model exceeds almost all the baselines in the Normal class and three overlapping patterns. Especially in SEO and EPO, the most common overlapping situations, our model achieves the highest performance, and the results are robust, which demonstrates the advantage of UniRel in processing overlapping triples. It can be seen that our model also makes improvements for almost all kinds of sentences regarding the number of triples. From simple situation (L $ = 1$) to complex case (L $ = 3$), UniRel still brings improvements, which shows the robustness of our model. In general, this experiment shows the power of our model in complex scenarios. We attribute the effectiveness to the captured rich interactions between entities and relations by the introduced Interaction Map, which is essential for solving the complex overlapping triple problem.

%% file: analysis.tex
\section{Analysis}

\subsection{Ablation Study}

In this section, we conduct ablation experiments on NYT and WebNLG datasets to study the effectiveness of the proposed Unified Representation and Unified Interaction as reported in Table \ref{tab:main_results}.

\subsubsection{Effect of the Unified Representation}

To study the effectiveness of Unified Representation, instead of assigning meaningful words for each relation, we use the placeholder \texttt{[unused]} of BERT to represent relations as marked as $\text{UniRel}_\textit{unused}$ in Table \ref{tab:main_results}.
Same as meaningless label ids, the embeddings of \texttt{[unused]} tokens are randomly initialized at the fine-tuning stage without augmenting the meaningful semantic information learned from pre-training. 
We can see a performance decay in terms of all the evaluation metrics on both datasets without Unified Representation, which indicates the importance of the semantic information for relational triple extraction.

We also notice the significant performance decrease of $\text{UniRel}_\textit{unused}$ on the WebNLG dataset.
We think it is because the WebNLG dataset has much fewer training data but defines far more relations compared to the NYT dataset. Such a contradiction makes many relations have few samples for training in the WebNLG dataset. 
And it is hard for a model to learn the deep semantic information with few examples from zero.

\begin{figure}[t]
    \centering
    \includegraphics[width=0.75\columnwidth]{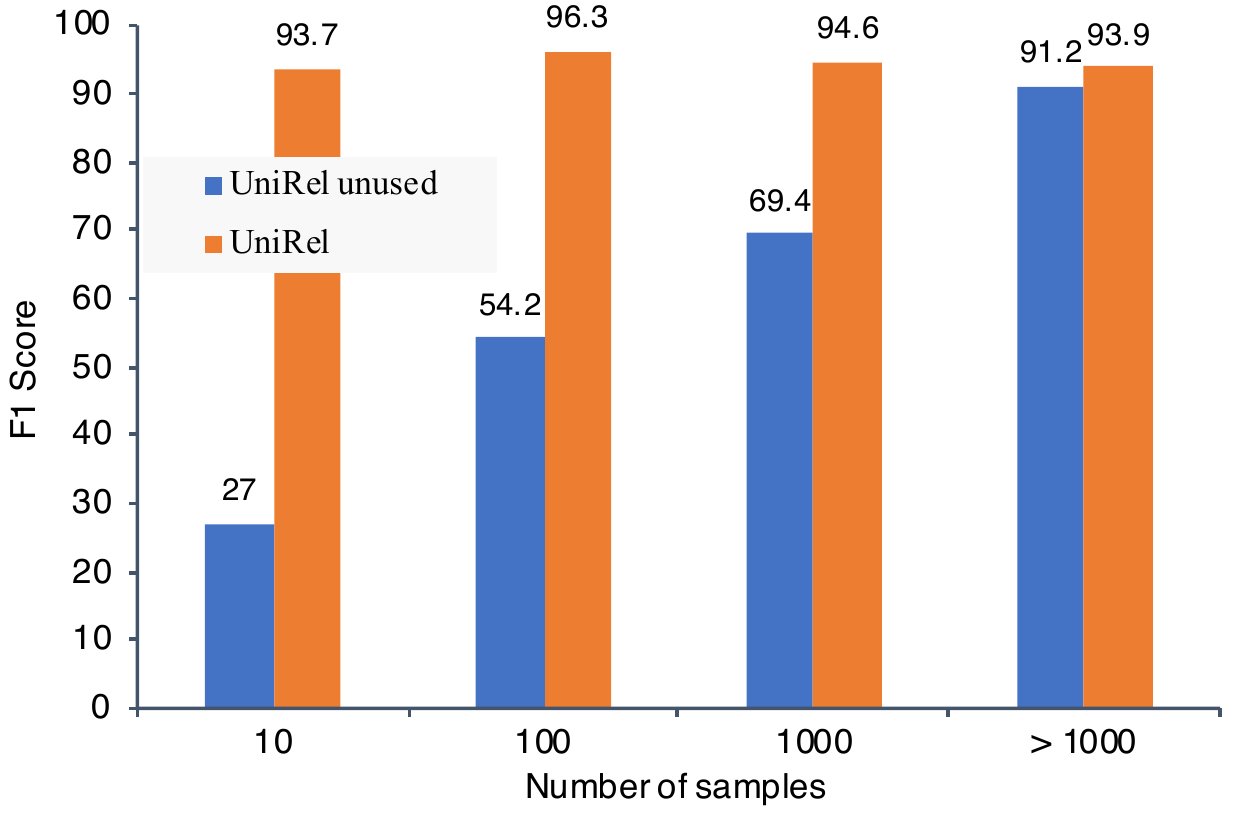} 
    \caption{F1-score on relations with different orders of magnitude samples in training set for UniRel (in Orange) and $\text{UniRel}_\textit{unused}$ (in Blue). 10/100/1000 means relations with less than or equal to 10/100/1000 samples.}
    \label{fig:instance_nums}
\end{figure}

To validate our assumption, we further analyze the performance of UniRel and $\text{UniRel}_{\textit{unused}}$ on relations with different orders of magnitude samples in training data on the WebNLG dataset.
As shown in Figure \ref{fig:instance_nums}, $\text{UniRel}_{\textit{unused}}$ performs well on relations with much samples ($\geq 1000$), but is limited when the number of samples decreases, which confirms our assumption.
In contrast, UniRel maintains a good performance level in the face of different numbers of samples. Especially with extremely few samples ($\leq 10$), UniRel performs at the same level as much samples ($ \geq 1000$) , which further demonstrates the effectiveness of Unified Representation.

\begin{figure}[t]
    \centering
    \includegraphics[width=1.0\columnwidth]{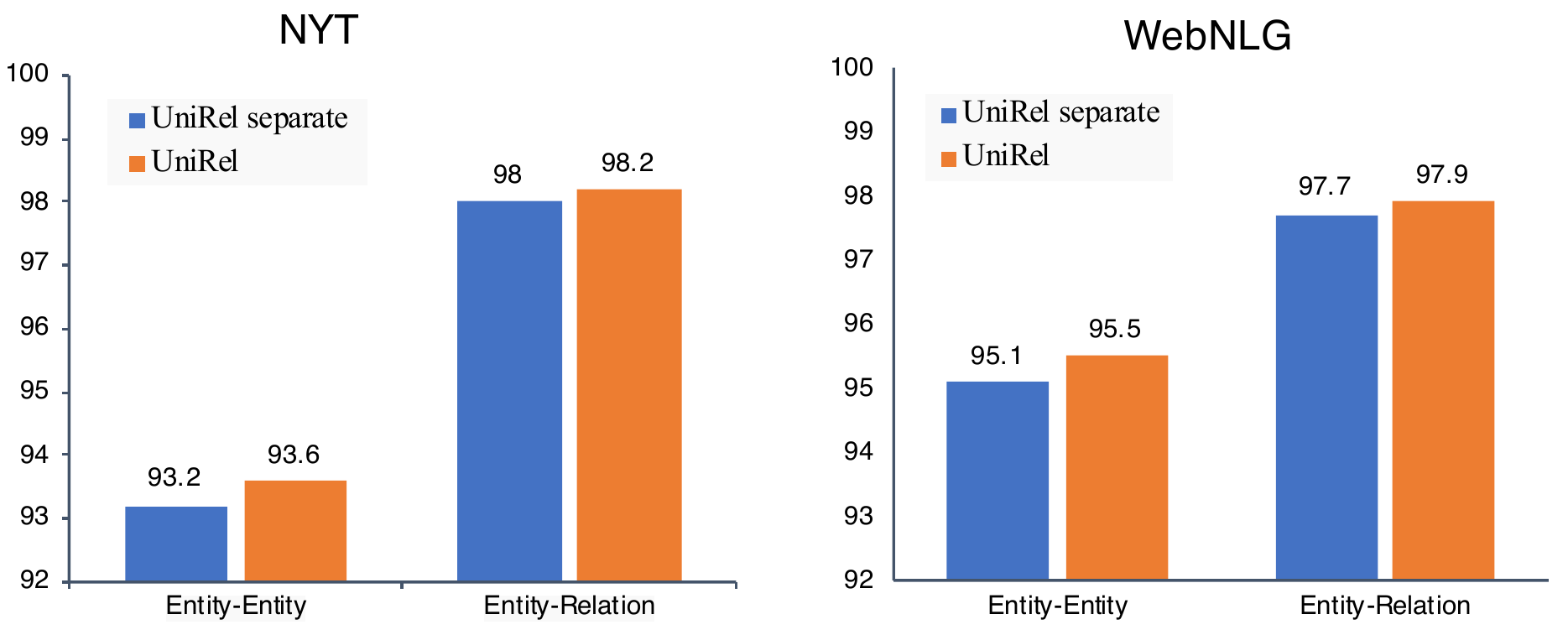} 
    \caption{F1-score on Entity-Entity Interaction and Entity-Relation Interaction for UniRel (in Orange) and $\text{UniRel}_{\textit{separate}}$ (in Blue).}
    \label{fig:EEandEP}
  \end{figure}

\begin{table}[h]
  \setlength\tabcolsep{4.8pt}
  \begin{tabular}{lccccc}
      \toprule
      \multirow{2}*{Model} &  \multicolumn{2}{c}{Training Time} & \multicolumn{2}{c}{Inference Time}\\
      \cmidrule(lr){2-3} \cmidrule(lr){3-5}
       &   NYT & WebNLG & NYT & WebNLG \\
      \midrule
      CasRel   & 1142 & 105 & 35 & 37 \\
      TPLinker & 2951 & 810 & 48 & 57 \\
      PRGC     & 3632 & 498 & 13 & 13 \\
      OneRel   & 2998 & 186 & 20 & 21 \\
      \midrule
      UniRel   & 967 & 119 & 12 & 14 \\
      \bottomrule
  \end{tabular}
  \caption{Computational Efficiency. Training time represents the time (second) needed to train one epoch. Inference time represents the average time (millisecond) to predict one sample.}
  \label{tab:computation}
\end{table}

\subsubsection{Effect of the Unified Interaction}

To study the influence of Unified Interaction, we take the relation sequence out of the input sentence to model the two kinds of interactions in a separate manner, and denoted as $\text{UniRel}_{\textit{separate}}$ in Table \ref{tab:main_results}.
Specifically, we first obtain the sequence embeddings of the input sentence and the natural language texts of relations severally with the same BERT encoder.
Then we apply two transform layers to get the query and key of the concatenation of the two embeddings.
Finally, the Interaction Map is outputted by doing dot production of query and key.

As $\text{UniRel}_{\textit{separate}}$ takes entity and relation as individual inputs, the deep Transformer network can only independently model the correlations inside entity pair and relation pair without capturing the interdependencies between entity-entity interactions and entity-relation interactions. As shown in Table \ref{tab:main_results},  $\text{UniRel}_{\textit{separate}}$ has marked performance degradation on both datasets compared to UniRel, which demonstrates the unified interaction's effectiveness.
We further analyze the performances of UniRel and $\text{UniRel}_{\textit{separate}}$ on different interaction types. As shown in Figure \ref{fig:EEandEP}, not only entity-relation, the F1-score of entity-entity also decreases without simultaneously modeling the interactions, which proves the interdependencies between the two kinds of interactions, and UniRel takes benefits from modeling them in a unified way.

\subsection{Computational Efficiency}

Table \ref{tab:computation} shows the comparison results of the computational efficiency between UniRel and some recent high-performance models. We report training and inference time on both NYT and WebNLG datasets. In this experiment, we follow previous works and set the batch size to 6/1 for training/inference. All the compared models are tested in the same hardware environment as declared in Section \ref{sec:de}.

UniRel shows a conspicuous computational efficiency performance in both training and inference time. Specifically, Compared to the SOTA model OneRel, on the NYT dataset, UniRel obtains 3$\times$ and 1.7$\times$ faster in the stage of training and inference, respectively. We think the reason is that UniRel $(O(N + M)^2)$ has a smaller prediction space than OneRel $(O(N \times M \times N ))$. Although CasRel performs similarly to ours regarding training time, UniRel obtains more than 2.6$\times$ speedups in the inference time. We attribute the efficiency to the design of the Interaction Map, which allows UniRel to narrow down the prediction space and directly leverage the off-the-shelf self-attention mechanism within the Transformer block.

\subsection{Visualization}
\begin{figure}[t]
  \centering
  \includegraphics[width=1.0\columnwidth]{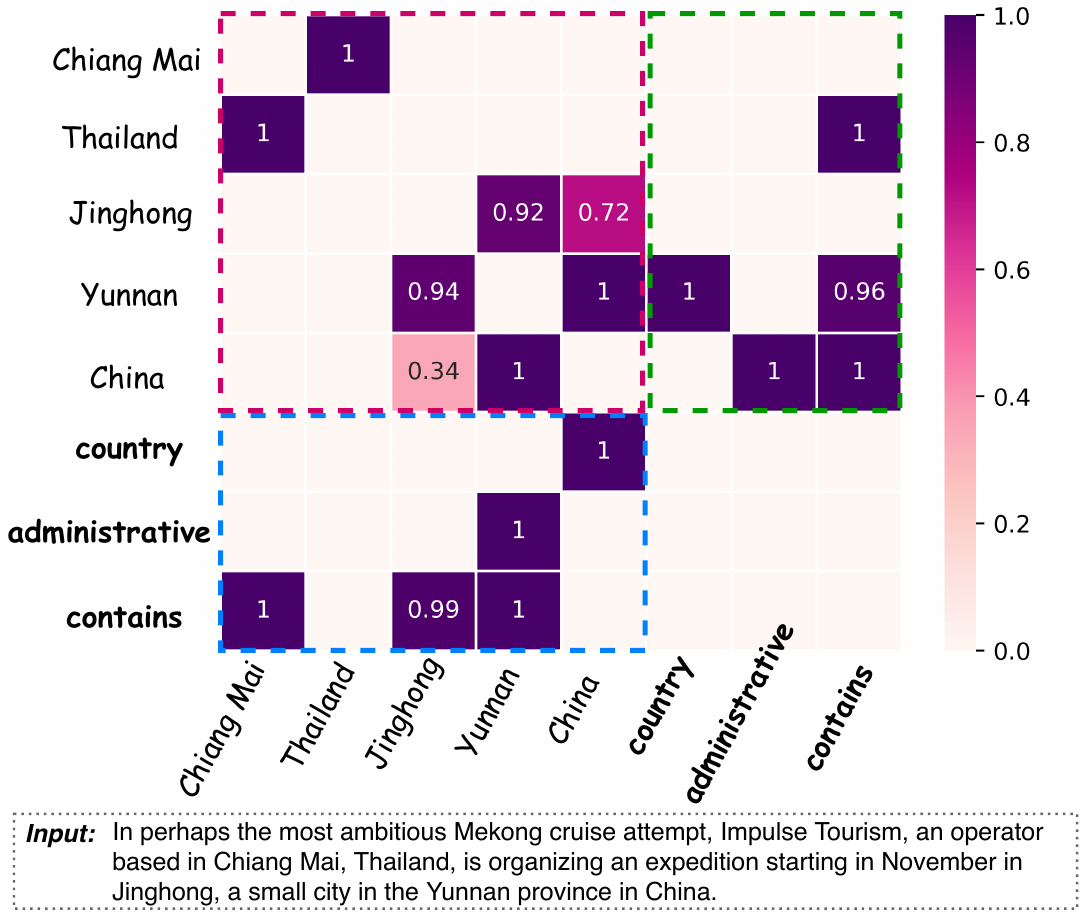} 
  \caption{Visualization of Interaction Map with input sentence sampled from NYT. Relations are in bold.}
  \label{fig:case_study}
\end{figure}

We visualize the Interaction Map to see how it works for relational triple extraction. As shown in Figure \ref{fig:case_study}, the red box represents the entity-entity interaction. The blue box and green box represent the entity-relation interaction for the subject and the object, respectively. From the map, we can extract all six relational triples: (\textit{Yunnan}, \texttt{country}, \textit{China}), (\textit{China}, \texttt{administrative\_divisions}, \textit{Yunnan}), (\textit{Thailand}, \texttt{contains}, \textit{Chiang Mai}), (\textit{Yunan}, \texttt{Contains}, \textit{Jinghong}), (\textit{China}, \texttt{Contains}, \textit{Jinghong}), and (\textit{China}, \texttt{Contains}, \textit{Yunnan}). 

\section{Conclusion}
In this work, we propose UniRel to fully leverage the rich correlations between entities and relations by resolving the heterogeneity.
Unified Representation eliminates the representation's heterogeneity by encoding both entity and relation into meaningful sequence embeddings.
Unified Interaction eliminates the interaction's heterogeneity by simultaneously modeling entity-entity interactions and entity-relation interactions in one single Interaction Map.
UniRel produces significant improvements over competitive baselines. We give a comprehensive analysis to further justify our design.

%% file: appendix.tex
\section{Supplemental Experiments}

\subsection{Influence of Random Seeds}
As shown in Table \ref{tab:5runs} , we conduct experiments with 5 different random seeds, and the results and their improvements are rather robust. We get an averaged performance of 93.62±0.07/94.52±0.12 on NYT/WebNLG with 5 runs in total, which robustly outperforms the previous SOTA methods.

\begin{table}[h]
    \begin{tabular}{lccccc}
        \toprule
        Dataset & 2019 & 2021 & 2022 & 2023 & 2024 \\
            
        \midrule
        NYT    & 93.6   & 93.7  & 93.6 & 93.7 & 93.5 \\
        WebNLG & 94.4   & 94.7  & 94.5 & 94.4 & 94.6 \\
        \bottomrule
    \end{tabular}
    \caption{F1-score of UniRel with different random seeds.}
    \label{tab:5runs}
\end{table}

\subsection{Influence of Relation Numbers}

\begin{figure}[t]
    \centering
    \includegraphics[width=1.0\columnwidth]{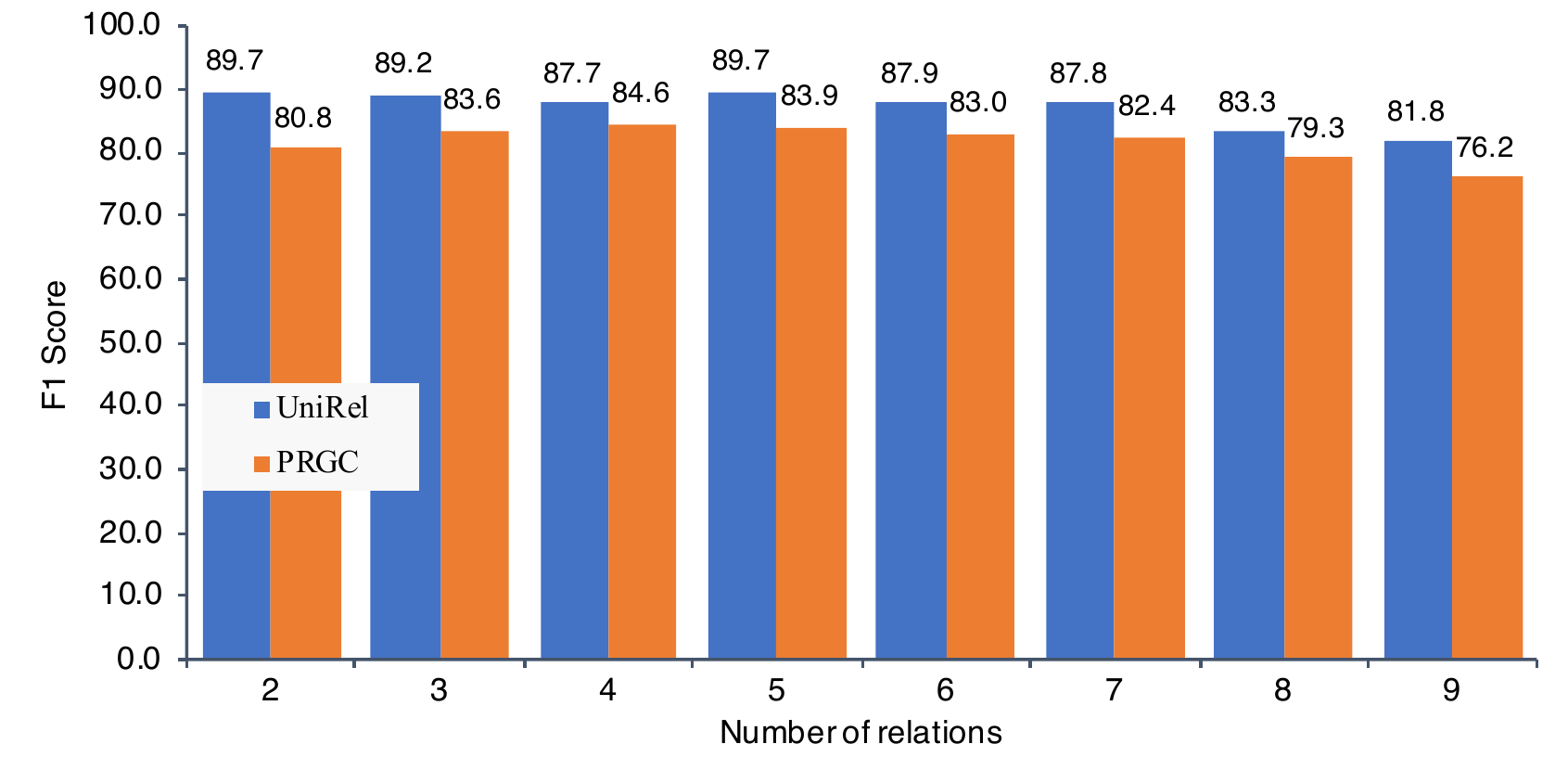} 
    \caption{F1-score w.r.t. different number of relations.}
    \label{fig:rel_numbers}
  \end{figure}

To study how the number of relations $M$ influences the performance of UniRel, as shown in Figure \ref{fig:rel_numbers}, we conduct ablation experiments with different $M$. We control the number of sentences of each relation in the range of [1500, 1700) to keep each relation has similar signals in each experiment.

From Figure \ref{fig:rel_numbers}, we can observe that UniRel has relatively stable improvements when the number of relations increases. For example, UniRel achieves both +5.6\% improvements when $M = 3$ and $M = 9$. The results show the effectiveness of UniRel is relatively stable with different numbers of relations.

\subsection{Relation-Word Mapping}

For clarity, we convert relations to natural language texts with human-picked words. The picked word is basically the most informative word within the label name that preserves its semantics, for example, ``founders'' for ``/business/company/founders''. We don't exploit any special selection strategy and need no formal annotation. For exceptional cases where the relation's labels are very similar, we simply resort to alternative words ("part” and "section”) or capitalization ("part” and "Part”) to make a distinction. This provides a semantic-aware initialization and the model will continue to optimize them. As a result, the performance of UniRel is very robust to choices of words. As shown in Table \ref{tab:relation-word mapping}, we conduct experiments with two different relation-word mappings of the NYT dataset, and the results show UniRel is robust to the mapping words.

\subsection{Extended to Multi-token Entity Setting}

With minimum adaptation, we can extend UniRel to the multi-token entity setting. The adaption processes are as follows:
1) repeat Interaction Map twice respectively for head and tail token of entity span to identify (subject-head, relation, object-head) and (subject-tail, relation, object-tail). 2) add a third Interaction Map between head-tail tokens to identify (head, relation, tail). 3) decode triples by linking the head and tail tokens of entities w.r.t. each relation.

As shown in Table \ref{tab:multi_token_setting}, UniRel achieves SOTA performance under the multi-token entity setting, which further indicates the effectiveness of the proposed methods.

\begin{table}[h]
    \renewcommand{\arraystretch}{1}
    \setlength\tabcolsep{3pt}
    \centering
    \begin{tabular}{lcccccc}
        \toprule
        \multirow{2}*{Model}      & \multicolumn{3}{c}{NYT} & \multicolumn{3}{c}{WebNLG} \\
        \cmidrule(lr){2-4} \cmidrule(lr){5-7} 
         & Prec.   & Rec.   & F1     & Prec. & Rec.  & F1       \\
        \midrule 
        NovelTagging    & 32.8    & 30.6   & 31.7   & 52.5    & 19.3  & 28.3  \\
        MultiHead  & 60.7    & 58.6   & 59.6   & 57.5    & 54.1  & 55.7  \\
        ETL-span  & 85.5    & 71.7   & 78.0   & 84.3    & 82.0  & 83.1  \\
        RSAN  & 85.7    & 83.6   & 84.6   & 80.5    & 83.8  & 82.1  \\        
        $\text{PRGC}_\textit{BERT}$  & 93.5    & 91.9   & 92.7   & 89.9    & 87.2  & 88.5  \\
        $\text{TPLinker}_\textit{BERT}$  & 91.4    & 92.6   & 92.0   & 88.9    & 84.5  & 86.7  \\        
        $ \text{GRTE}_\textit{BERT}$  & 93.4 & \textbf{93.5} & \textbf{93.4} & \textbf{92.3} & 87.9 & 90.0 \\
        $\text{OneRel}_\textit{BERT}$  & 93.2 & 92.6 & 92.9 & 91.8 & 90.3 & 91.0 \\
        \midrule                          
        $\text{UniRel}$  &\textbf{ 93.7} & 93.2 & \textbf{93.4} & 91.8 & \textbf{90.5} & \textbf{91.1} \\
        \bottomrule
    \end{tabular}
    \caption{Results of the multi-token entity setting. The highest scores are in bold.}
    \label{tab:multi_token_setting}
\end{table}

\begin{table*}[h]
    \begin{tabular}{lcc}
        \toprule
        Relation & Mapping A & Mapping B \\
            
        \midrule
        /business/company/advisors    & advisors   & counselor   \\
        /business/company/founders    & founders   & creator   \\
        /business/company/industry    & industry   & sector   \\
        /business/company/major\_shareholders    &  holding  &  shareholder  \\
        /business/company/place\_founded    &  founded  &  establish \\
        /business/company\_shareholder/major\_shareholder\_of    & shareholder   & holder   \\
        /business/person/company & company & corporation \\
        /location/administrative\_division/country & country & state \\
        /location/country/administrative\_divisions & administrative & administration \\
        /location/country/capital & capital & Capital \\
        /location/location/contains & contains & include \\
        /location/neighborhood/neighborhood\_of & neighbor & neighbour \\
        /people/deceased\_person/place\_of\_death & death & Death \\
        /people/ethnicity/geographic\_distribution & geographic & Geographic \\
        /people/ethnicity/people & people & People \\
        /people/person/children & children & Children \\
        /people/person/ethnicity & ethnicity & ethnic \\
        /people/person/nationality & nationality & national \\
        /people/person/place\_lived & lived & live \\
        /people/person/place\_of\_birth & birthplace & birth \\
        /people/person/profession & profession & career \\
        /people/person/religion & religion & Religion \\
        /sports/sports\_team/location & location & Location \\
        /sports/sports\_team\_location/teams & teams & Teams \\
        \midrule
        Prec./Recall/F1. & 93.5/94.0/93.7 & 93.9/93.4/93.6 \\
        \bottomrule
    \end{tabular}
    \caption{Two different relation-word mappings for NYT}
    \label{tab:relation-word mapping}
\end{table*}